*Editors*
Dragi Kocev, Nikola Simidjievski, Ana Kostovska,
Ivica Dimitrovski & Žiga Kokalj

# DISCOVER THE MYSTERIES OF THE MAYA

Selected Contributions from
the Machine Learning Challenge &
the Discovery Challenge Workshop, ECML PKDD 2021

# Preface

Remote sensing has greatly accelerated traditional archaeological landscape surveys in the forested regions of the ancient Maya. Typical exploration and discovery attempts, besides focusing on whole ancient cities, focus also on individual buildings and structures. Recently, there have been recent successful attempts of utilizing machine learning for identifying ancient Maya settlements. These attempts, while relevant, focus on narrow areas and rely on high-quality aerial laser scanning (ALS) data which covers only a fraction of the region where the ancient Maya were once settled. Satellite image data, on the other hand, produced by the European Space Agency's (ESA) Sentinel missions, is abundant and, more importantly, publicly available.

In particular, the Sentinel-1 satellites are equipped with Synthetic Aperture Radar (SAR) operating globally with frequent revisit periods, while the Sentinel-2 satellites are equipped with one of the most sophisticated optical sensors (MSI and SWIR), capturing imagery from visible to medium-infrared spectrum with a spatial resolution of 10-60m. While the latter has been shown to lead to accurate performance on a variety of remote sensing tasks, the data from the optical sensors is heavily dependent on the presence of cloud cover, therefore combining it with radar data from the Sentinel-1 satellites provides an additional benefit. Integrating Sentinel data has been shown to lead to improved performance for different tasks of land-use and land-cover classification.

Motivated by this, we organized a machine learning competition to locate and identify ancient Maya architectures (buildings, aguadas, and platforms) by performing integrated image segmentation of different types of satellite imagery (from Sentinel-1 and Sentinel-2) data and aerial laser scanning data (lidar data). The main area of interest for the challenge is a region around Chactún, Mexico - one of the largest known Maya urban centers located in the central lowlands of the Yucatan peninsula. The area is characterized by low hills with constructions and surrounding seasonal wetlands (bajos). Chactún is located in the northern sector of the depopulated Calakmul Biosphere Reserve in Campeche, Mexico, and is completely covered by tropical semi-deciduous forest. Its urban core, composed of three concentrations of monumental architecture, has a number of plazas surrounded by temple pyramids, massive palace-like buildings, and two ball courts. Ceramics collected from the ground surface, the architectural characteristics, and dated monuments, indicate that the center started to thrive in the Preclassic period, reaching its climax during the Late Classic (ca. A.D. 600–1000), and had an important role in the regional political hierarchy.

The challenge dataset consists of tiles derived from Sentinel-1, Sentinel-2, and ALS (lidar) data, with associated annotation masks:

- *Sentinel-1 data*: Level-1 Ground Range Detected (GRD) products of IW acquisition mode were acquired, for ascending and descending orbit, with two polarizations (VV and VH), and Sigma0 as backscatter coefficient. Values of the backscatter coefficient were converted to decibels (dB), fitted to [-30, 5] dB interval, and normalized to the interval [0, 1]. Then, multiple temporal statistics were calculated for each tile: mean, median, standard deviation, coefficient of variance, 5th, and 95th percentile, pixel-wise for each year separately (2017, 2018, 2019, 2020) and for the whole period (2017-2020). Each Sentinel-1 TIFF file consists of 120 bands (5 x 60 bands; 24 by 24 pixels; float).
- *Sentinel-2 data*: Level-2A products were acquired with reflectance data from 12 spectral bands (B01, B02, B03, B04, B05, B06, B07, B08, B8A, B09, B11, B12). All bands were resampled to 10-meter resolution. Due to the geographical and climate characteristics of the test area in the central Yucatan peninsula (frequent small convective clouds or haze), a cloud mask was calculated for each acquisition date. Acquisition dates with cloud cover above 5% were excluded. There are 17 valid acquisition dates in the period from 2017-2020 with 12 spectral bands and a cloud mask for each date. In total, each TIFF file, therefore, consists of 221 bands (17 x 13 bands; 24 by 24 pixels; float).

- *ALS (lidar) data*: ALS data is provided in the form of a visualization composite consisting of sky-view factor (band 1), positive openness (band 2), and slope (band 3) in separate bands. The tiles coincide with Sentinel tiles and therefore have 480 by 480 pixels (3 bands, 8-bit).
- *Annotation masks*: separate masks for buildings, platforms, and agendas (480 by 480 pixels, 8-bit, 0 feature present, 255 not present). A valid *output of the models* is a prediction of the segmentation masks for each tile in the test set. In particular, for each tile, one has three masks, one for each of the classes of man-made structures – buildings, platforms, and aguadas.

Each tile measures 240 x 240 meters and has a spatial resolution of 10 meters for Sentinel data and 0.5 meters for ALS data. The Sentinel-1 and Sentinel-2 data for each tile are stored separately in multi-band TIFF files.

We received solutions from 25 teams that actively took part in the competition, with more than 100 requests for downloading the data. The submissions were evaluated using standard measures for estimating the quality of image segmentation methods. In particular, the predicted segmentation masks were compared to the ground-truth masks using *Intersection Over Union (IoU)* score. The IoU score also referred to as critical success, evaluates the overlap between the predicted segmentation mask and the ground truth, or in other words the ratio of correctly predicted regions among the predicted regions. The submissions included the prediction of the segmentation masks for each tile in the test set. For each tile, the solutions were required to include three masks, one for each of the classes of structures – buildings, platforms, and agendas. Each submission was evaluated using the average IoU score between the submitted predictions and the ground-truth masks. More specifically, each submission was evaluated with 4 different average Intersection Over Union (IoU) scores, one for each class of structures and one computed on all predictions. The winning solutions were determined using the overall average IoU score.

The final leaderboard[1] comprised solutions that differ in their modeling approaches, data modalities being used as well as data preprocessing techniques. In particular, almost all of the top 7 solutions leveraged only lidar data, while the 5th place solution combined also Sentinel 2. The top-performing competitors build their solutions around variants of DeepLabV3 and UneXt50 with ResNet100 encoders and/or HRNETs, with the top 3 solutions combing separate models, learned for each mask separately, into ensembles leading to better predictive performance.

**Table 1** Top 7 performing solutions from the competition.

| # | User | Avg. IOU | Avg. IOU of aguadas | Avg. IOU of platforms | Avg. IOU of buildings |
|---|---|---|---|---|---|
| 1 | Aksell | 0.8341 | 0.9844 | 0.7651 | 0.753 |
| 2 | ArchAI | 0.8316 | 0.9873 | 0.7611 | 0.7464 |
| 3 | German Computer Archaeologists | 0.8275 | 0.9851 | 0.7404 | 0.7569 |
| 4 | dmitrykonovalov | 0.8262 | 0.9836 | 0.7542 | 0.7409 |
| 5 | The Sentinels | 0.8183 | 0.9854 | 0.73 | 0.7394 |
| 6 | taka | 0.8127 | 0.9771 | 0.7354 | 0.7256 |
| 7 | cayala | 0.811 | 0.9863 | 0.7082 | 0.7386 |
| 7 | sankovalev | 0.811 | 0.9844 | 0.7421 | 0.7066 |

---

[1] available at `https://biasvariancelabs.github.io/maya_challenge/`

Driven by the great interest in the competition as well as its uniqueness, in terms of the area of application of machine learning, we invited several top solutions to a workshop, to further discuss motivations, challenges, and experiences during the challenge. The workshop was organized as part of the Discovery Challenge Track of The European Conference on Machine Learning and Principles and Practice of Knowledge Discovery in Databases (ECML PKDD 2021)[2]. Four invited teams responded positively and prepared reports discussing their solutions. These proceedings include the reports on the solutions provided by the following teams:

- Chapter 1: Team **ArchAI** – Matthew Painter and Iris Kramer, 'Discover the Mysteries of the Maya';
- Chapter 2: Team **German Computer Archaeologists** – Jürgen Landauer, Burkhard Hoppenstedt, and Johannes Allgaier, 'Image Segmentation To Locate Ancient Maya Architectures Using Deep Learning';
- Chapter 3: Team **The Sentinels** – Thorben Hellweg, Stefan Oehmcke, Ankit Kariryaa, Fabian Gieseke, and Christian Igel, 'Ensemble Learning for Semantic Segmentation of Ancient Maya Architectures';
- Chapter 4: Team **cayala** – Christian Ayala, Carlos Aranda, and Mikel Galar, 'A Deep Learning Approach to Ancient Maya Architectures Detection Using Aerial Laser Scanning Data'.

## Organizers

Nikola Simidjievski, Bias Variance Labs, Slovenia, University of Cambridge, UK & Jožef Stefan Institute, Slovenia
Dragi Kocev, Bias Variance Labs & Jožef Stefan Institute, Slovenia
Ana Kostovska, Bias Variance Labs & Jožef Stefan Institute, Slovenia
Ivica Dimitrovski, Bias Variance Labs, Slovenia
Žiga Kokalj, ZRC SAZU, Slovenia
Bertrand Le Saux, Phi-Lab, ESRIN, European Space Agency, Italy
Andrej Draksler, ZRC SAZU, Slovenia
Maja Somrak, ZRC SAZU, Slovenia
Tatjana Veljanovski, ZRC SAZU, Slovenia
Ivan Kitanovski, Bias Variance Labs, Slovenia
Tomaž Stepišnik, Bias Variance Labs & Jožef Stefan Institute, Slovenia
Matej Petković, Bias Variance Labs & Jožef Stefan Institute, Slovenia
Panče Panov, Bias Variance Labs & Jožef Stefan Institute, Slovenia
Sašo Džeroski, Jožef Stefan Institute, Slovenia
Sara Aparicio, Phi-Lab, ESRIN, European Space Agency, Italy
Sveinung Loekken, Phi-Lab, ESRIN, European Space Agency, Italy

The organizing institutions of the competition were:
Bias Variance Labs, Slovenia
Research Centre of the Slovenian Academy of Sciences and Arts (ZRC SAZU), Slovenia
Jožef Stefan Institute, Slovenia
Phi-Lab, ESRIN, European Space Agency, Italy

## Discovery Challenge Chairs, ECML PKDD 2021

Paula Brito, Universidade do Porto, Portugal
Dino Ienco, INRAE, UMR TETIS, University Montpellier, France

---

[2] https://2021.ecmlpkdd.org/

## Acknowledgments

Acknowledgment of data collection and preparation: ALS data acquisition was financed by the KJJ Charitable Foundation, presided by Ken and Julie Jones (USA) and The Chactún Regional Project: Study of an Archaeological Landscape in the Central Maya Lowlands (ARRS project J6-7085; 2016-2018). Data processing was further financed by research programs Anthropological and Spatial Studies (ARRS P6-0079; 2015-2021) and Earth Observation and Geoinformatics (ARRS P2-0406; 2019-2024) and by the project AiTLAS:Artificial Intelligence Toolbox for Earth Observation (cont.4000128994/19/D/AH, ESA, 2020-2021).

The organization of the competition "Discover the mysteries of the Maya" was supported by a grant from the European Space Agency (ESRIN): AiTLAS - Artificial Intelligence toolbox for Earth Observation (ESA RFP/3-16371/19/I-NB).

July 2022                                                            *Nikola Simidjievski*
*Ivica Dimitrovski*
*Ana Kostovska*
*Žiga Kokalj*
*Dragi Kocev*

# Contents

# Chapter 1
# Discover the Mysteries of the Maya

Matthew Painter[1] and Iris Kramer[1,2]

## 1.1 Introduction

Discovering new archaeological sites is key to better understand history and to protect our heritage. Manual assessment of large amounts of remote sensing data for the detection of unknown archaeological sites is very time consuming. In the Discover the Mysteries of the Maya challenge competition, participants were tasked to create an approach to automatically identify archaeological sites on LiDAR and Satellite Imagery. This paper presents our second place solution, for which we have focused on the latest deep learning advancements. Our solution was driven by robust validation rather than reliance on leaderboard scores. Indeed, our validation allowed us to submit less than 20 unique submissions for leaderboard evaluation, over the whole competition.

## 1.2 Data

At the start of the challenge we examined the data, considering class distributions, rough sizes of class instances, and found that the classes were extremely unbalanced (see Fig. 1.1 'Original'). Class imbalance can result in poor predictive performance, specifically for the minority class. There are a few means to deal with class imbalance, and we initially chose to sample the data such that each class (including background) was equally represented (Fig. 1.1 'Equal'). Eventually we settled on the weighted distribution in Fig. 1.1 ('Ours') which massively de-emphasised background and slightly de-emphasised aguadas.

We also found that there was very limited number of training images for a deep learning solution. We therefore focused our approach to include pretraining, augmentation and techniques to artificially increase training data.

For our solution, we considered using both the LiDAR and Sentinel data. However, at visual inspection we were unable to identify the objects on the Sentinel imagery because of the forest canopy and relatively low resolution of the imagery (10m). Neither were we able to train our deep learning models to pick out the structures. We briefly explored super-resolution of the Sentinel data, given that we had images of the same sites over multiple dates. This however did not prove beneficial. We therefore only used LiDAR (0.5m) in our final solution.

---

ArchAI LTD, London E16 2DQ, UK
e-mail: iris@archai.io
http://www.archai.io · University of Southampton, Southampton SO17 1BJ, UK

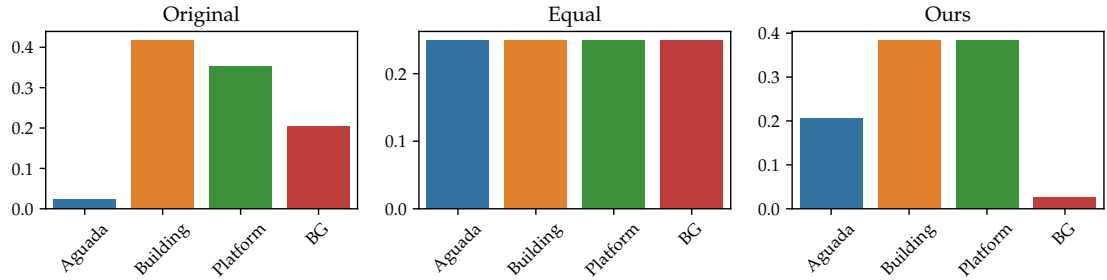

**Fig. 1.1** Estimates of the proportion of each class in the dataset (Original), the equally weighted variant (Equal) and our final, weighted sample dataset (Ours).

## 1.3 Training

We trained all models using PyTorch [13], the Adam optimiser [8], and a combination (sum) of the Dice and Focal [11] losses. Learning rate was initially $10^{-4}$ with some runs reducing the learning rate by a factor of 10 for the last 10 epochs, and others also performing a linear learning rate warmup over the first 5 epochs. Most experiments were ran for 100 epochs, with a batch size of 12 or 20. Specific details will be provided when discussing the final submission.

Robust Validation

Good validation techniques are vitally important for challenges such as this, since it is extremely easy to over-fit to public leaderboards. We utilised a 5-fold cross validation to avoid this, since we found it sufficiently traded-off experiment runtime and robust performance estimation. The 5-fold validation also had the advantage of allowing us to ensemble the folds to make leaderboard submissions, a technique we found improved results dramatically.

Ensembling

One of our biggest gains in accuracy was brought about with ensembling. This allows small mistakes of different models to be averaged out across the predictions. We found best performance by using a variety of architectures instead of using the highest scoring models from the same architecture. This is because different architecture generally make different mistakes.

Augmentation

Since the training dataset consisted of only around 1800 images, we ran all solutions using standard data augmentations applied probabilistically, implemented using Kornia [14]. Specifically, we applied horizontal flipping, vertical flipping, random rotation ($[-\frac{\pi}{2}, \frac{\pi}{2}]$), random translation (up to 15% of image size) and random scale change ($[0.5, 1.5]$), all applied with probability 0.5. We explored additional augmentations (Gaussian noise, colour jitter) but found they hurt validation performance.

Pseudo Labelling

Pseudo-labelling [10] is a common technique (particularly in hidden leaderboard challenges) to increase the available training data. It consists of using a well trained model to predict labels for the test data, which can then be merged with the training data to form a larger (if slightly noisy) dataset. We chose to perform pseudo-labelling using an ensemble that performed well in our validation results. Pseudo labels were used for some models in our final solution, and provided a large improvement in validation scores.

## 1.4 Model Architectures

We explored a number of architectures for the challenge, and we will detail important findings here.

Baseline

The baseline for the challenge was a fine tuned DeepLabV3 [2] using torchvision (COCO [12]) pretrained weights with a ResNet101 [7] backbone. When ensembling the 5 validation folds, the baseline performed well, however its architecture is not well suited to the problem. Since the buildings and platforms have well defined boundaries, it is important to preserve high resolution information. DeepLabV3 computes features at a relatively low resolution level, and simply upsamples them for the final prediction. This works well for large structures, but fails to properly segment smaller structures.

DeepLabV3+

Subsequent to DeepLabV3, DeepLabV3+ [3] was introduced which added a simple decoder architecture to the DeepLabV3 model. This allows the model to utilise the low resolution features (and the context generated by the ASPP module) as well as higher resolution features introduced to the decoder through skip connections (conceptually similar to UNet [15] skip connections). We found that this dramatically improved performance on buildings and platforms, which constituted the majority of the small structures in the dataset. Similar to the baseline, the DeepLabV3+ for our experiments had a ResNet101 backbone, however it was pretrained on CityScapes [4], rather than COCO.

HRNet

HRNet [17] was used by the winning solution of SpaceNet 7 challenge which is one of the most recent remote sensing challenges [19]. HRNet merges information from each scale into all others, and maintains both high and low resolution representations throughout the network. In theory, by mitigating the downsampling present in DeepLabV3+, segmentation of the small structures should improve further. Motivated by the SpaceNet challenge result and the higher resolution features compared to DeepLabV3+, we implemented HRNet using pretrained ImageNet weights [20].

For this challenge, we adapted HRNet by doubling the resolution of the network, we call this HHRNet. HHRNet was the highest performing single model on our local validation scores, however it did not improve over our best ensemble of DeepLabV3+ models on the public leaderboard. In the absence of ensembling, our validation scores suggest that HHRNet would generalise best to new data. Indeed, in the SpaceNet challenge, HRNet with suitable post-processing outperformed all other solutions, many of which comprised of large ensembles.

## 1.5 Final Solution

Our final solution was an ensemble of 5 models and their associated cross validation folds. All models used our weighted sampling method, the Dice Focal loss, and a learning rate drop by a factor of 10 in the last 10 epochs. Specifically, the ensemble composition is given in Table 1.1, where we filtered out poor performing (by validation) folds from the final ensemble.

Table 1.1 Final solution ensemble

|              | DeepLabV3 | DeepLabV3+ | HHRNet | DeepLabV3+ | HHRNet |
|--------------|-----------|------------|--------|------------|--------|
| Epochs       | 60        | 100        | 100    | 100        | 100    |
| LR Warmup    | 0         | 5          | 5      | 5          | 5      |
| Pseudo-Labels| ×         | ×          | ×      | ✓          | ✓      |
| N Folds      | 4         | 4          | 5      | 4          | 4      |

## 1.6 Potential Improvements

Post Processing or Distance Loss Penalties

By visually reviewing our resulting masks we noticed that we mostly lost accuracy at the boundaries of objects. We considered improving the outcome with techniques that encourage precise boundaries. We commonly found confusion between buildings and platforms. In the SpaceNet challenges this problem is often solved by using multi-channel masks [19] which denote object interiors, edges, and contacts between objects. Other methods to improve results include post-processing, refinement networks [21] and distance loss penalties [1].

Mixed Sample Data Augmentations

Since we saw a large (validation) performance increase with pseudo-labelling additional data, we would liked to have explored data augmentations which produce plausible new data. Mixed sample data augmentation methods such as FMix [6], CutMix [22] or perhaps MixUp [23] might provide (probably small) additional improvements.

External Datasets or Synthetic Data

A common method to improve performance is to use external data as a pretraining step, ideally with the same modality and task as the target data. We expect that pretraining our models on archaeological earthworks elsewhere in the world would have improved performance. Such benchmarks are rare and have only limited data, although there are examples [9]. Similarly, alternative networks could be used that were pretrained on LiDAR datasets for different tasks such as Lunar LiDAR [16]. However, if such datasets are not openly available they can be very costly to create. A final potential avenue to increase training data is through generative modelling or manual synthetic data creation. Whilst it is difficult to generate data which accurately matches the true dataset, it can be effective as a pretraining step before fine tuning onto the real data [18].

Alternate Weighted Sampling

Per class pixel count distributions differed from the image-wise count. Aguadas were by far the largest structure, but appeared in the fewest images. Buildings were significantly smaller on average than platforms, but appeared in more images. Our solution performed best on aguadas, then platforms, then buildings, an ordering which matches the average sizes, from largest to smallest. Potentially, weighting based on the *expected pixel sizes* rather than our custom weighting would have been a better option.

## 1.7 Conclusions

Detecting archaeological earthworks on LiDAR data is a challenging problem, progress in which can have a high impact on heritage understanding and protection. By using the latest advancements in deep learning, our team was able to significantly improve the baseline. Our major improvements to the baseline will be contributed to the AiTLAS toolbox [5] which should aide efforts to improve automated detection of archaeology. By digging into the literature we also found several potential improvements which we hope will help the field move towards higher accuracy in the future. For our team, consisting of archaeologists and machine learning experts, this competition has been a great milestone and we will continue to contribute to the detection of archaeological sites across the globe.

## Acknowledgements

The authors would like to thank the competition organisers and sponsors for their great contribution to the field of archaeology. Our participation in the challenge was supported by grants from the Royal Academy of Engineering (under the Enterprise Fellowships scheme), the SPRINT Programme (through the UK Space Agency) and the Geovation Accelerator Programme from Ordnance Survey.

# Chapter 2
# Image Segmentation To Locate Ancient Maya Architectures Using Deep Learning

Jürgen Landauer[1], Burkhard Hoppenstedt[2], and Johannes Allgaier[3]

**Abstract** Exploration of the Maya forest region remotely through machine learning has recently accelerated. Using experts to manually look at satellite data is time-consuming and expensive. The machine learning competition Discover the mysteries of the Maya addresses this problem and calls for a competition to improve the performance of state-of-the-art models to automatically detect objects of interest using satellite images. With a given LiDAR image, the model should detect three classes of objects: Aguadas, buildings and platforms. We have set up a pipeline that essentially consists of three steps. First, we generate synthetic data in three different ways to increase the training set. In the second step, we mix them with the real training data and then train an ensemble of DeepLabV3+ and HRnet networks. In the third step, we applied thresholds to improve the segmentation masks. We achieved an average intersection over union (IOU) of 0.8275 for all three classes and the best score of 0.7569 for the building class.

**Key words:** Deep Learning, Image Segmentation, Computer Vision, Synthetic Data

## 2.1 Introduction

Archaeologists face the challenge of finding ancient buildings in very large areas. In this context, the assistance of automated object detection in images may help research to improve their speed and detection accuracy. This paper refers to the data set of the Maya challenge and its task to locate ancient Maya architectures (with the classes *auguadas*, *buildings* and *platforms*) through binary image segmentation. For a given input, the model should segment the image into `class` or `background` pixels. Each input results in three binary masks (one per class) for the output as shown in Figure 2.1. Note that class pixels are not necessarily disjoint, as buildings may be on top of a platform.

Landauer Research, Ludwigsburg, Germany
e-mail: juergenlandauer@gmx.de · University of Ulm,
Institute of Databases and Information Systems, Germany
e-mail: burkhard.hoppenstedt@uni-ulm.de · University of Würzburg,
Institute of Clinical Epidemiology and Biometry, Würzburg, Germany
e-mail: johannes.allgaier@uni-wuerzburg.de

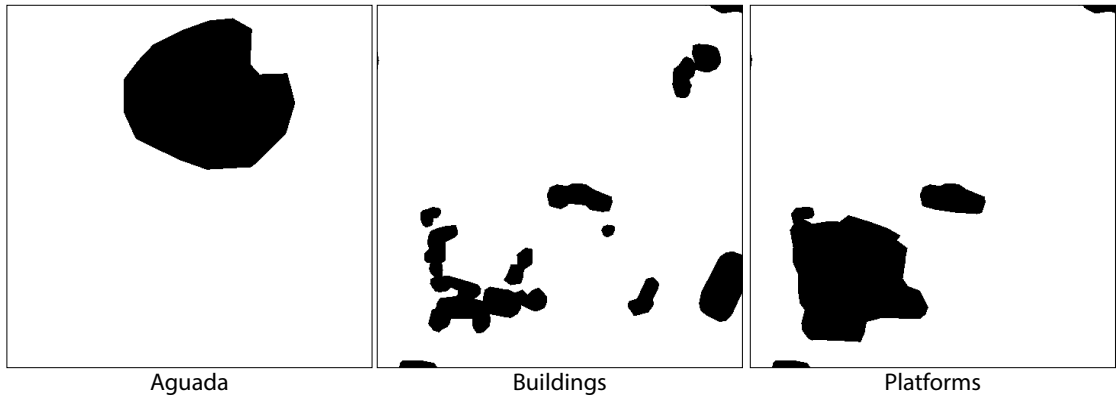

**Fig. 2.1** Samples of label data for each class. The training data consists of 64 images of aguadas, 1129 images of buildings, and 951 images of platforms. The samples are taken from three different inputs.

## 2.2 Methods

In this section, we give a brief overview of the Machine Learning pipeline to obtain our results. The machine learning pipeline can be divided into the three phases *pre-processing*, *Deep Learning model training* and *post-processing*.

### 2.2.1 Pre-Processing

For the improvement of the model, we considered it necessary to increase the amount of training data. Since the possibilities of data augmentation techniques from given frameworks were not specific enough, we decided to generate synthesized data using Python's `NumPy` library. We therefore implemented three different approaches resulting in three additional datasets, which were included in varying degrees into the training data, depending on their performance for different network architectures (see Fig. 2.3):

1. `rectangular-cropped`: Compute a rectangular bounding box of object of interest, e.g. an *aguada*
2. `pixel-precise-cropped`: Compute a precise border line of object
3. `padded`: same as 2. but with $N$ added padding pixels around the object to get a smooth transition between class and non-class pixels.

The pipeline looks as follows: We split the training images into two groups: Either images showing a class (*auguada, building, platform*) or empty images without any class. From images showing a class, we copy the pixels of interest in one of the three styles (`rectangular-cropped, pixel-precise or padded`). Next, for each empty image (ref. `Target_Origin`), we randomly copy a cropped image (`Class_Slice`) and paste it at a random location in the empty image. We store the information where the class image was inserted and generate a mask (`Generated_Mask`) from it. For details of the process, see Fig. 2.2.

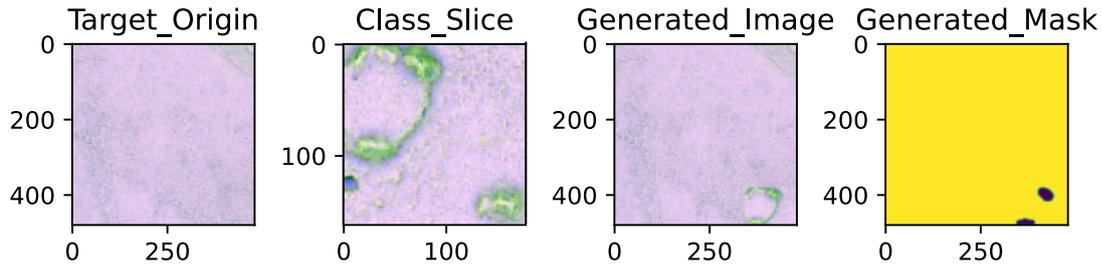

**Fig. 2.2** Three Steps of our image generation process. In the first step, we loop over all the empty images without any class (`Target_Origin`), then randomly drag from an image with class pixels, cut out the class pixels there in one of the three styles (`Class_Slice`), and then place them at a random position in the target image (`Generated_Image`). Finally, we save the new, associated mask (`Generated_Mask`).

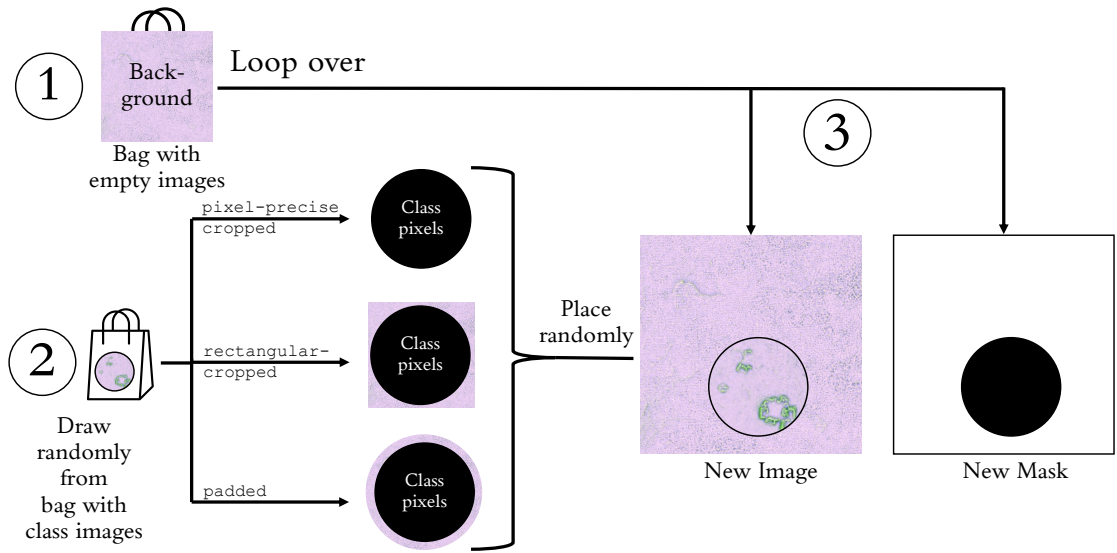

**Fig. 2.3** Image generation pipeline using three different techniques `rectangular-cropped, pixel-precise-cropped and padded`. The three steps are are marked in circles. In (1), we loop over all images containing only background, in (2), we draw randomly from a list that contains images from the class we want to generate. Finally, in (3), we place a cropped version of that class image at a random position in the empty background images. The rose background is a real sample of a `.tif` image.

### 2.2.2 Deep Learning model training

For each of the three archaeological object classes two neural network architectures were trained: (1) a DeepLabV3+[1] with a ResNet-101 backbone, and (2) an HRNet[6] with a HRNet-W30 backbone. To obtain optimal results, the LiDAR training dataset, including the synthetic data, was randomly shuffled and then split into 5 folds of equal size using cross validation, which iteratively serve as validation dataset (and the remaining 4 folds are used as the training dataset), resulting into a 80/20 (training/validation) data split. This training resulted in 30 models (3 classes x 2 architectures x 5 folds). For each class we ensembled the corresponding 10 models during prediction. In essence, we computed the (unweighted) mean of all 10 predictions for each pixel and stored it in a grayscale image. The image then corresponds to the output of the last layer or the ensembled network (i.e. the probability or confidence score) and serves as input for the post-processing step outlined in the next section. In our opinion, additional key success components of our solution were: (A) The use of the fast.ai library with its simple but effective interface on top of Pytorch[3], (B) Data augmentation techniques: Even with synthetic images, the share of pixels containing the classes to be detected was relatively small. So we oversampled the minority classes, e.g. samples containing the `aguada` class were simply duplicated six times during training. In addition, CutMix data augmentation[8] helped for some classes. (C) Progressive resizing: Each model was trained twice, once with half size and once with full size images. This helped models regularize better and reduced training time. (D) Focal Foss as loss function usually works well for imbalanced data and was used here, too[4]. (E) Mixed precision training: Unlike other frameworks, the fast.ai library can invoke Nvidia CUDA for training with either 16 or 32-bit float numbers. We always trained with 16 bit except for the last training run. This not only saves time but also improves model performance through better regularization (as suggested in [2] ). (F) Modern training loop: we utilized these improvements over other training frameworks (inspired by L. Wright's blogpost[7]): Replacing the Adam optimizer with Ranger and replacing the One-Cycle training with a Flat+Cosine annealing training cycle.

### 2.2.3 Post-Processing

After the deep learning based segmentation process, we apply two different kinds of filters to receive the final binary result. The output of the previous phase consists of probability values for each pixel. If these probability values are scaled up to the range [0,255], they can be seen as greyscale images. For these images, we introduce a threshold, which we refer here to as *probability threshold* (see Table 2.4). The lower the threshold, the more likely it is for a pixel to be accepted as a class pixel. It is evident that by being less strict in accepting pixel as class pixel, the number of false positives will increase. Therefore, the probability threshold is combined with another threshold, which we denote as *blob threshold*. After the binarization of the greyscale image using the probability threshold, we detect all objects in the image and calculate their area in pixels. We filter small areas, as we assume a minimal size for the objects to be detected. Moreover, the area around the detected object did not have a high probability value, otherwise it would have been accepted as class pixel while lowering the probability threshold. Therefore, the small blob is not likely to belong to a real object. The combination of the two introduced thresholds improved our pipeline significantly.

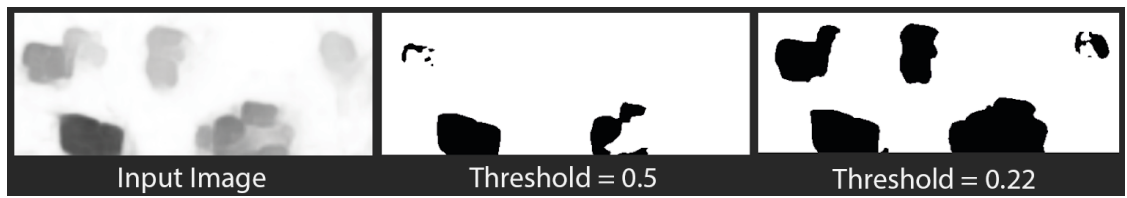

**Fig. 2.4** Effects of probability threshold. A probability threshold of 0.50 means that the pixel is classified as `class` if the model surer than 49.8 % that this pixel is a `class` pixel. A probability filter of 21.57 % means the model must be surer this value.

## 2.3 Results

In the competition we scored a third place with IoU values of 0.9851 (Aguada), 0.7404 (Platforms), 0.7569 (Buildings). This results in a overall IoU of 0.8275 (see Table 2.1). The code is available on GitHub.

**Table 2.1** Final Leaderboard of the competition with our team (German Computer Archaeologists, GCA) on the third place. The column values show the Intersection of Union (IOU) for the object class and all classes on average. For each column, the highest score is marked bold. In total, 25 teams competed.

| Rank | Team | Total | Aguadas | Platforms | Buildings |
|---|---|---|---|---|---|
| 1 | Aksell | **0.8341** | 0.9844 | **0.7651** | 0.7530 |
| 2 | ArchAI | 0.8316 | **0.9873** | 0.7611 | 0.7464 |
| **3** | **GCA** | 0.8275 | 0.9851 | 0.7404 | **0.7569** |
| 4 | dmitrykonovalov | 0.8262 | 0.9836 | 0.7542 | 0.7409 |
| 5 | The Sentinels | 0.8183 | 0.9854 | 0.7300 | 0.7394 |

## 2.4 Discussion

We tried to find out what information the *Sentinel data* contained for the segmentation task at hand. The first problem we saw was the low resolution (24x24) of the input images. In particular, i.e. the `building` class has very granular grids for segmentation, so it was not feasible for us to scale the 24x24 input to 480x480 in a meaningful way. Also, a data analysis showed that the two sets of sentinel data contributed approximately 37 % of the total amount of data due to their smaller resolution (See table below). Moreover, the bands within a sample have a high redundancy because they show the same location at different times. That is, the overall information contribution is much lower. Hence, we did not include the satellite datasets into our processing pipeline. The LiDAR data consists of three color channels, which in fact are three different visualizations (i.e. transformations) derived from the same raw LiDAR data. This inevitably causes some information (and hence detection quality) loss compared to using the raw data, as we have outlined in our own research on LiDAR in Deep Learning[5] but fortunately the quality was excellent, and definitely good enough for the instance segmentation task at hand.

**Table 2.2** Gross information contents of data sources

|  | Bands | PpS[*] | BpP[**] | Type | Bits | Share |
|---|---|---|---|---|---|---|
| **Sentinel 1** | 120 | 69 120 | 32 | float | 2 211 840 | 25 % |
| **Sentinel 2** | 221 | 127 296 | 8 | uint | 1 018 368 | 12 % |
| **Lidar** | 3 | 230 400 | 24 | uint | 5 529 600 | 63 % |

[*]PpS = Pixel per sample, [**]BpP = Bits per Pixel

Concerning the *Pre-Processing*, each data generation types results in different effects on each data class. The classes do not benefit equally from the generated images. Unfortunately, we cannot recommend a preferred setup. Furthermore, we struggled to distinguish between the classes platform and building. The labeled data for these classes seems often to be similar. We assume, that another labeling might improve the segmentation accuracy, e.g. the introduction of a new class building on a platform. Finally, In some cases, we noticed inconsistencies in the granularity of the labeled data (see Figure 2.5). Especially for the class building, where structures can be very filigree this might cause a large difference in the segmentation accuracy.

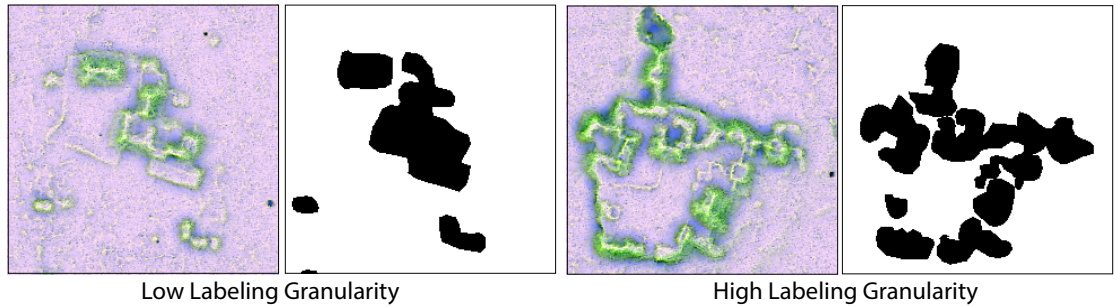

Low Labeling Granularity　　　　　　　　　High Labeling Granularity

**Fig. 2.5** Label data for two LiDAR images and their corresponding masks. Especially the resolution of the contours varies between the masks.

# Chapter 3
# Ensemble Learning for Semantic Segmentation of Ancient Maya Architectures

Thorben Hellweg[1], Stefan Oehmcke[2], Ankit Kariryaa[2], Fabian Gieseke[1,2], and Christian Igel[2]

**Abstract** Deep learning methods hold great promise for the automatic analysis of large-scale remote sensing data in archaeological research. Here, we present a robust approach to locating ancient Maya architectures (buildings, aguadas, and platforms) based on integrated segmentation of satellite imagery and aerial laser scanning data. Deep learning models with different architectures and loss functions were trained and combined to form an ensemble for pixel-wise classification. We applied both training data augmentation as well as test-time augmentation and performed morphological cleaning in the post-processing phase. Our approach was evaluated in the context of the "Discover the mysteries of the Maya: An Integrated Image Segmentation Challenge" at ECML PKDD 2021 and achieved one of the best results with an average IoU of 0.8183.

**Key words:** Deep Learning, U-Net, DeepLabv3, Remote Sensing, Satellite Data, Lidar Data, Maya Archaeology

## 3.1 Introduction

Machine learning-based segmentation of aerial and satellite imagery is a promising tool for archaeological research. As areas of interest are often difficult to access or hidden by vegetation, satellite or aerial imagery offer the possibility to survey large territories at low cost. However, manual segmentation by experts is time-consuming and expensive in terms of human labor costs.

Recently, deep learning-based approaches such as the U-Net [11] have successfully been employed for detecting objects in remote sensing. In special domains such as archaeology, however, only a small number of annotated training samples are usually available. Additionally, ground truth masks may exhibit labelling noise reflecting annotator-related biases. Against this background, the "Discover the mysteries of the Maya: An Integrated Image Segmentation Challenge"[1] at ECML PKDD 2021 sought for contributions on the localisation of three classes of ancient Maya architectures (buildings, aguadas[2], and platforms) by

---

Department of Information Systems, University of Münster
e-mail: \{thorben.hellweg,fabian.gieseke\}@uni-muenster.de · Department of Computer Science, University of Copenhagen
e-mail: \{stefan.oehmcke,ak,fabian.gieseke,igel\}@di.ku.dk

[1] https://biasvariancelabs.github.io/maya_challenge/comp/
[2] artificial rainwater reservoirs [13]

performing integrated image segmentation of different types of satellite imagery and aerial laser scanning data.

In this work, we present our submission to the challenge, which is based on a heterogeneous ensemble of deep learning models.[3] In an effort to find a suitable architecture for the segmentation task, models based on the U-Net and DeepLabv3 architectures were trained and extended. As not all data provided were equally suitable for the segmentation tasks, training inputs were selected from airborne laser scanning (ALS) and Sentinel-2 data. To increase robustness and to improve accuracy, we applied training data augmentation as well as test-time augmentation and used morphological cleaning as a final post-processing step. Since we eventually opted not to use a validation set in favour of larger training data (and thus supposedly better models), we combined the fitted models into an ensemble to mitigate the effect of overfitting.

Our approach generated a robust segmentation of all Maya structures and achieved an average IoU that was 5 percentage points higher compared to the provided DeepLabv3 challenge baseline.

### 3.2 Ancient Maya Settlements Dataset

We applied our approach to the challenge dataset covering the Chactún archaeological site [2, 7, 13]. The dataset comprises tiles derived from Sentinel-1, Sentinel-2, and ALS data, along with associated annotation masks. For Sentinel-1 and Sentinel-2, the dataset includes statistics and imagery for the years 2017 – 2020. Each Sentinel-1 TIFF file consists of 120 bands ($24 \times 24$ pixels) and contain several temporal statistics, such as mean, median, standard deviation, coefficient of variance, for each year separately and for the entire period. The Sentinel-2 data groups 12 spectral bands (B01 – B12), each resampled to a 10 m resolution. For the period 2017 – 2020, there are 17 recordings with cloud masks and cloud cover below 6%. In total, each TIFF file therefore consists of 221 bands ($17 \times 13$ bands; $24 \times 24$ pixels). Contrary to the low-resolution Sentinel tiles, the ALS data is provided with a resolution of 0.5 m per pixel in form of a visualization composite consisting of sky-view factor (band 1), positive openness (band 2), and slope (band 3) in separate bands ($1 \times 3$ bands; $480 \times 480$ pixels) [13].

The data is split into training and testing data. For the training set, annotation masks are provided, separately for buildings, platforms, and aguadas ($480 \times 480$ pixels, 8-bit, 0 corresponds to the target being present, 255 to not present). The distribution of Maya structures in the training set is highly imbalanced. While 1211 of the 1765 tiles contain one of the three archaeological structures, only 64 of these contain the structure aguada. In comparison, there are 952 tiles with platforms and 1129 tiles with buildings. Buildings and platforms are also often found together within a tile. Furthermore, buildings are typically placed on platforms, but can also be found outside of platforms.

### 3.3 Learning Segmentation Models

#### 3.3.1 Pre-Processing

As the resolution of the Sentinel data is at least twenty times lower than that of the ALS data, much of the provided Sentinel data is potentially of little use for training the models. For this reason, we decided to only use the bands 2, 3, 4 (BGR), and band 8 (NIR) of the Sentinel-2 data for training. To reduce the Sentinel-2 images, we created median composites of cloud-free pixels. Both ALS and Sentinel-2 data were min-max normalised to a range between 0 and 1. For Sentinel-2, we used the 90% quantile value instead of the maximum value to create a more robust scaling. Due to the low resolution, no data from Sentinel-1 were used to train the models. We initially split the available training set into two folds (training/validation), containing 80% and 20% of the samples, respectively. After testing the validity of our approach, we finally trained different models on the entire training set, based on which the ensemble is formed.

---

[3] The code is available at `https://github.com/thllwg/maya-challenge-public`.

**Table 3.1** Considered augmentation schemes

| method | data | parameters | probability |
|---|---|---|---|
| **basic:** | | | |
| random cropping | all | between 256 and 400 pixel of ALS size | 100% |
| vertical flip | all | | 50% |
| horizontal flip | all | | 50% |
| **advanced:** | | | |
| random rotation | all | between 0 and 359 degrees | 25% |
| Gaussian blur | ALS | with kernel size 11 and randomly between 0.1 and 2 | 25% |
| additive noise | all inputs | on each input individually | 25% |
| uniform | | between 0 and 0.1 | 50% |
| normal | | standard deviation of 0.03 | 50% |

We also applied data augmentations to further mitigate the problem of the imbalanced and (relatively) small dataset. We divided our augmentations into normal and "advanced" augmentations. In contrast to the advanced augmentations, the normal augmentations – such as rotation and flipping – do not change the statistics of the images or drastically disturb the images, and are therefore considered safe to be applied. Depending on the model trained, only normal or both normal and advanced augmentations were used. Each augmentation has an individual probability of being applied, see Table 3.1.

### 3.3.2 Architectures

We experimented with different model architectures and combined them to obtain an ensemble of heterogeneous segmentation models. In particular, we made use of modified versions of the U-Net [11] architecture, which has already been successfully used for similar segmentation tasks in archaeological research [2]. In addition, we considered variants of the DeepLabv3 [3] architecture.

We first investigated several encoder networks to replace the downsampling part of the original U-Net. Although successfully employed in other research applications, using ResNet [5] as our encoder did not improve the model performance and was eventually omitted in favour of MNasNet, a comparatively small network optimized for mobile devices [15], and Swin-B, a self-attention network with pre-trained weights [9]. We used different activation functions for each: ReLU in ResNet, ELU [4] in MNasNet, and GELU [6] in Swin-B. For the Deeplabv3 architecture [3], we considered pre-trained weights for the ResNet101 encoder [5]. The pre-trained encoders were not optimized for the initial epochs to give the decoder time to adjust to the pre-trained weights.

For the upsampling path in the decoder part of the U-Net, instead of using bi-linear interpolation to initialise added pixels with the weighted average surrounding pixels, sub-pixel convolution with ICNR initialisation (pixel shuffle) was used [1, 12]. Contrary to most encoders that halve the resolution at each block, Swin-B downsamples to a quarter of the resolution per block. Consequently, we upsample two times in a row, which showed better results than directly upsampling to four times the resolution. As many super-resolution methods are known to generate artifacts, we applied $2 \times 2$ averaging filters after each upsampling operation to reduce artifacts [14]. Inspired by Zhang *et al.* [17], we introduced a self-attention layer in the second last upsampling layer, which allows to better model long-range dependencies and which also increases the receptive field. As batch normalization in the decoder reduced the model performance, we replaced it with spectral normalization [17].

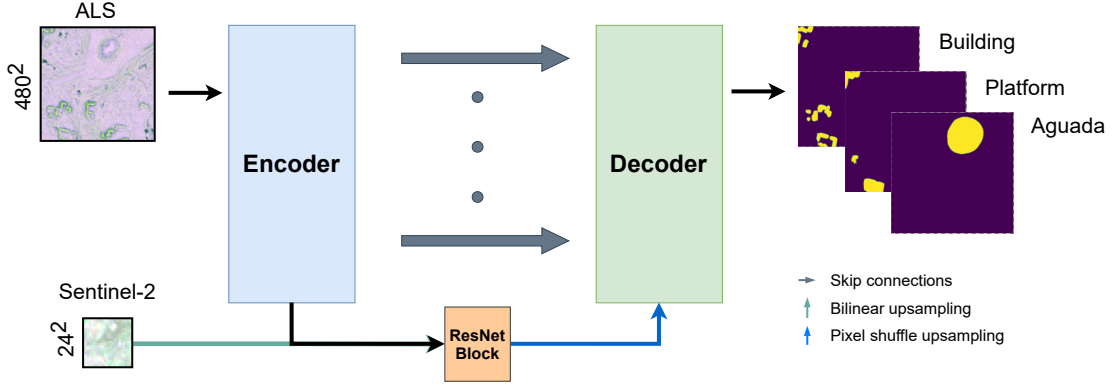

**Fig. 3.1** An overview of our U-Net architecture

**Table 3.2** Overview of different parameter assignments for: learning rate (lr), use of warm-restarts and cycling learning rates (cycle), batch size (BS), effective batch size (EBS), excluding aguada (no ag.), number of model checkpoints (# cp), usage of self-attention layer, and random crop size (rcs).

| Architecture | Encoder | Loss | lr | cycle | BS | EBS | no ag. | # cp | SA | rcs |
|---|---|---|---|---|---|---|---|---|---|---|
| U-Net | Swin-B | BCE | $6 \cdot 10^{-5}$ | Yes | 52 | 104 | No | 3 | Yes | 256 |
| | | BCE | $6 \cdot 10^{-5}$ | Yes | 52 | 208 | No | 4 | Yes | 256 |
| | | BCE | $6 \cdot 10^{-5}$ | No | 52 | 208 | No | 1 | Yes | 256 |
| | | VGG | $6 \cdot 10^{-5}$ | No | 32 | 96 | No | 1 | Yes | 256 |
| | | BCE | $3 \cdot 10^{-4}$ | No | 30 | 90 | No | 1 | No | 400 |
| | MNasNet | VGG | $6 \cdot 10^{-5}$ | No | 6 | 12 | No | 1 | Yes | 256 |
| | | BCE | $6 \cdot 10^{-5}$ | No | 52 | 104 | No | 1 | Yes | 256 |
| | | BCE | $3 \cdot 10^{-4}$ | No | 6 | 6 | No | 1 | No | 400 |
| DeepLabv3 | Resnet101 | BCE | $3 \cdot 10^{-5}$ | No | 16 | 64 | Yes | 1 | No | 400 |
| | | BCE | $3 \cdot 10^{-5}$ | No | 16 | 16 | No | 4 | No | 400 |

We conducted several experiments to incorporate the Sentinel-2 data at different stages of the U-Net encoder. Eventually, we decided to add a final cross-over ResNet block layer after the encoder, which received the concatenated encoder output and Sentinel-2 data as input. The Sentinel-2 data was upscaled to the same resolution as the encoder feature map.

### 3.3.3 Training

Various loss functions were used to train the networks. We found that good results could be consistently achieved with binary cross-entropy (BCE). (Focal) Tversky loss, which initially seemed promising considering the imbalance of the data, did not converge and was not used in the final ensemble for any of the models. In an effort to add a spatial loss function, we used the VGG loss [8] together with BCE. The VGG loss is calculated by inputting the predicted and true segmentation into an ImageNet pre-trained VGG16 model and by backpropagating the smooth L1 difference of the intermediate layers (blocks 2, 3, and 4). We used the Adam optimizer and considered various learning rates in the range between $1 \times 10^{-2}$ and $1 \times 10^{-6}$. To further smooth the learning curve during training, we accumulated the gradient over multiple batches (effective batch size). We used dropout and cosine learning rates with warm restarts. The different parameters assignments are shown in Table 3.2.

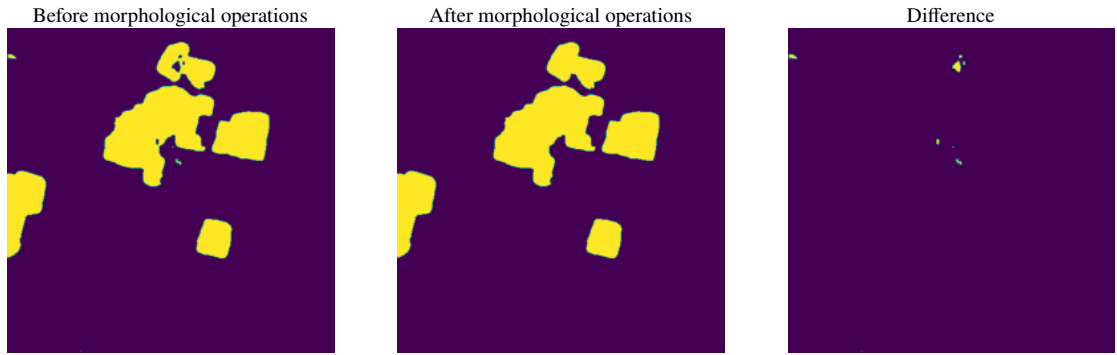

**Fig. 3.2** Effect of morphological cleaning on the platforms predicted in tile 1832: A small platform on the top left is removed and the holes in the platforms are filled.

Instead of training individual models for each of the structures to be predicted, we used multi-task learning to exploit information shared among two or more connected tasks. In doing so, we assumed that predictions of platforms and buildings in particular are closely linked. With one exception, all models were trained to jointly predict aguadas, platforms, and buildings. For one model, we excluded aguada, which means that we did not consider the aguada mask during training as well as prediction. While oversampling of aguadas increased our aguada IoU at the beginning, its influence later became negligible and was, therefore, discontinued. When we stopped validating our models using a holdout set, we could no longer reliably determine whether models were already overfitting during training. In addition, since different models previously evaluated exhibited different strengths, the models trained on the full dataset were combined into an ensemble. Our final ensemble consisted of ten model configurations with 18 model instances (note that some ensemble members were represented at multiple points via checkpoints, where the validation error was lowest).

### 3.3.4 Post-Processing

To acquire the predictions of a single model, we applied test-time augmentation, where we flipped and rotated (in 90° steps) each test image. Hence, due to these operations and their combinations, several predictions were obtained per test image. The final output was the average of these predictions after reversing the applied operations. This was done to make our predictions more robust towards rotations and partly translations.

The final ensemble was obtained via soft majority voting of all chosen members to counter overfitting and to smooth the prediction borders. We also tried hard voting, where, instead of using the probabilities, only the discrete class decisions are used. The soft voting strategy produced better results and was, hence, preferred. Due to the time constraints, we could not investigate the effect of stacking [10, 16], which can potentially improve results and which can be used to identify models that perform better at predicting certain classes.

We noticed that, at times, our models predicted buildings, platforms, or aguadas, which were only a couple of pixels wide. Therefore, based upon the size distribution of the buildings, platforms, and aguada in the training masks, we removed the predicted objects whose size was below a certain threshold value. We used a different threshold value when the objects were partially or completely on the boundary of an image. However, the thresholds were chosen ad-hoc and could be improved in the future. Likewise, very small objects (at the edge) could already be removed from the training masks during pre-processing.

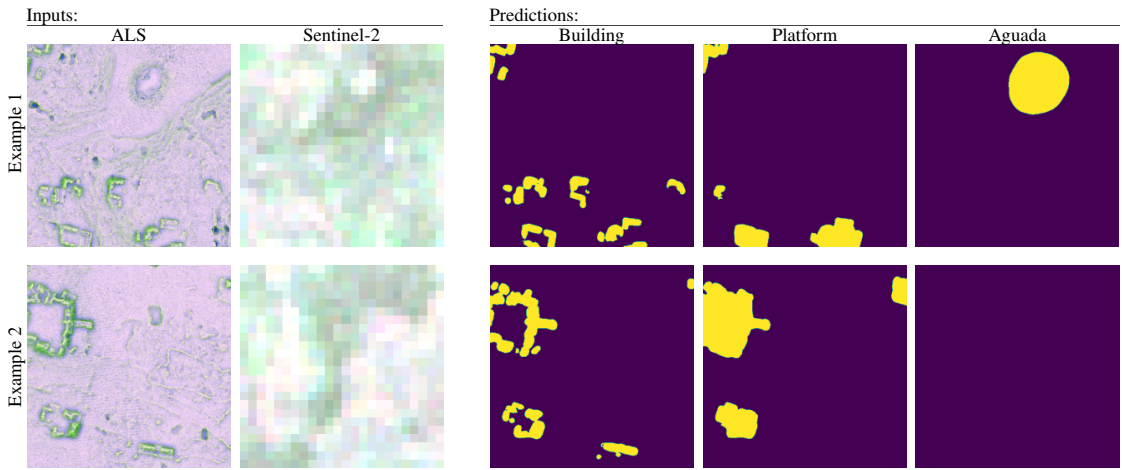

**Fig. 3.3** Prediction of the model ensemble on two sample tiles.

The predicted masks had, at times, holes inside of predicted objects. In the training masks, we observed that the aguada and platform masks were usually convex without any gaps, and while buildings could take arbitrary shapes, they did not have holes either. Therefore, we decided to fill the holes in the predicted polygons during post-processing. Figure 3.2 demonstrates the effect of the morphological cleaning on the platforms predicted in tile 1832.

## 3.4 Results

The key components of our approach (i.e., ensembling heterogeneous models, augmentation in both training and testing, and morphological cleaning) turned out to be important drivers to improve the overall quality. Test-time augmentation helped to eliminate overconfident incorrect predictions. While it mainly affected the edges of the predicted masks, entire object masks could be corrected as well. By combining the individual models into an ensemble, this effect was further enforced. As the number of ensemble members increased, so did our average IoU. At the same time, however, the effect of test-time augmentation decreased, as overconfident predictions were corrected by ensembling many model instances.

Although only Sentinel-2 bands with the highest native resolution were selected for training, our models did not make use of the Sentinel-2 inputs (providing non-matching Sentinel-2 inputs after training did not lead to changes in the predictions).

An iterative evaluation of intermediate results on the test set via the challenge competition platform allowed for a best-breed approach, in which the best masks for each Maya structure were combined for the final contribution. We have refrained from improving our final submission in this way.

Figure 3.3 exemplifies the performance of our final model. It shows the predicted buildings, platforms and, aguadas for two sample tiles along with the corresponding ALS and Sentinel images. The organisers of the challenge provided the following results on the test set. Our ensemble approach was among the best submissions and achieved an average IoU of 0.8183 (aguadas: 0.9854, platforms: 0.7300, buildings: 0.7394).

**Acknowledgements**

SO, AK, FG, and CI acknowledge support by the Villum Foundation through the project *Deep Learning and Remote Sensing for Unlocking Global Ecosystem Resource Dynamics (DeReEco)*.

# Chapter 4
# A Deep Learning Approach to Ancient Maya Architectures Detection Using Aerial Laser Scanning Data*

Christian Ayala[1], Carlos Aranda[1], and Mikel Galar[2]

**Abstract** Traditionally, archaeologists have manually inspected Airborne Laser Scanning (ALS) data to detect ancient Maya structures, resulting in a costly and time-consuming process. ALS data is expensive to capture and prone to poor resolution and quality. Therefore, in the last decade, promising automation approaches combining ALS data with deep learning models have emerged. Additionally, in the last few years, high-resolution satellite imagery has been considered to further improve the generalization capabilities of these models, since large study areas are not typically covered by ALS data. This work presents the 7th solution for the "Discover the mysteries of the Maya" Challenge. The aim of the challenge is to locate ancient Maya structures making use of deep learning techniques. The proposed solution applies binary decomposition strategies to reduce the complexity of the multi-class semantic segmentation problem, taking a different approach for each sub-problem.

**Key words:** Maya archaeology, Convolutional neural network, Semantic segmentation, LiDAR, Binary decomposition.

## 4.1 Introduction

The manual inspection of Airborne Laser Scanning (ALS) [9] is a time-consuming process traditionally performed by archaeologist. Raw ALS data is often converted to Digital Elevation Models (DEMs) to help experts with this task [12]. However, the great deal of manual intervention required has a negative effect on the data analysis workflow.

In the last decade, deep learning-based techniques have been successfully applied to almost every application domain, ranging from computer vision [4] to natural language processing [3]. Convolutional Neural Networks (CNNs) have become the de facto solution for computer vision and image processing tasks, including semantic segmentation.

Tracasa Instrumental, Calle Cabárceno 6, 31621 Sarriguren, Spain; e-mail: `cayala@itracasa.es` · Institute of Smart Cities (ISC), Public University of Navarre (UPNA), Arrosadia Campus, 31006 Pamplona, Spain; e-mail: `mikel.galar@unavarra.es`

* Christian Ayala was partially supported by the Government of Navarra under the industrial PhD program 2020 reference 0011-1408-2020-000008. Mikel Galar was partially supported by Tracasa Instrumental S.L. under projects OTRI 2018-901-073, OTRI 2019-901-091 and OTRI 2020-901-050, and by the Spanish MICIN (PID2019-108392GB-I00 / AEI / 10.13039/501100011033).

CNNs have been also applied to a wide range of remote sensing use cases such as the detection of building footprints [7] and the extraction of road networks [2]. Moreover, there are a great deal of works which combine ALS data with CNNs for archaeological purposes [20, 21, 22]. Therefore, our hypothesis is that CNNs can be effectively applied to ALS data to detect ancient Maya sctructures hidden under the thick forest canopy. Accordingly, this work opens up new possibilities for a wide range of archaeological applications, since costs and human effort are drastically reduced.

The proposed approach decomposes the multi-class semantic segmentation problem into multiple (a priori simpler) binary ones. This outlook reduces the complexity of the multi-class problem, making it possible to design different CNNs architectures to each class.

The remainder of this article is organized as follows. The challenge is outlined in Section 4.2. Then, Section 4.3 presents the proposed solution. Thereafter, Section 4.4 disscuss the results obtained. Finally, Section 4.5 concludes this work and present some future research.

## 4.2 Challenge description

### 4.2.1 Data-set

The data-set [20] belongs to the "Discover the mysteries of the Maya" Challenge and is located in Chactún, Mexico, as it is shown in Figure 4.1. This area, which is completely covered by tropical semi-deciduous forest, is characterized by low hills with constructions and surrounding seasonal wetlands.

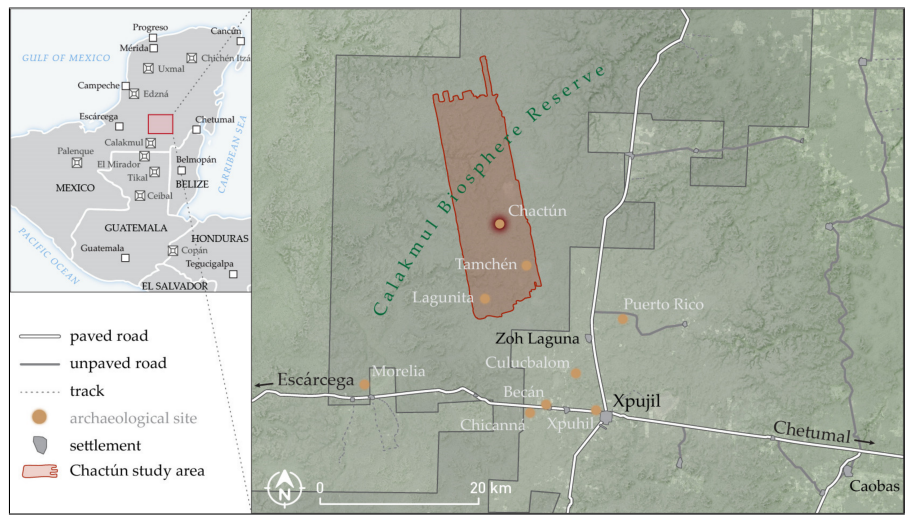

**Fig. 4.1** Location of the scanned area in the Calakmul Biosphere Reserve, Campeche, Mexico. [20]

The data-set combines Sentinel-1 (S1) and Sentinel-2 (S2) high-resolution satellite imagery (at 10 m) with ALS data (at 0.5 m). Regarding the S1 imagery, the Level-1 Ground Range Detected (GRD) products in Interferometric Wide (IW) swath mode for both ascending and descending orbits have been considered. Even though four polarization modes (VV, VH, HH, HV) are provided by S1, only the dual vertical polarization (VV, VH) is used in this data-set. Accordingly, backscattering coefficients have been computed using this data and converted to decibels (dB). Moreover, after fitting the $[-30, 5]$ dB interval they have finally been normalized to [0, 1] dB. Finally, multiple temporal pixel-wise statistics have been calculated for each tile: mean, median, standard deviation, coefficient of variance, 5th, and 95th percentile, not only for each year separately (2017, 2018, 2019, 2020) but also for the whole period (2017-2020). As a result, each 24 × 24 pixels S1 tiles consists of 120 bands (6 *statistics* × 2 *polarization modes* × 2 *orbits* × 5 *years*).

Regarding the S2 imagery, the Level-2A bottom of atmosphere (BOA) reflectance products have been considered. All the spectral bands (B01 – B12) excluding B10, have been stacked after resampling the ones with lower resolution to 10 m. Additionally, due to the geographical and climate characteristics of the study area, a cloud mask has been calculated for each acquisition date. Moreover, dates with a cloud cover percentage greater than 5 have been excluded, resulting in 17 valid dates in the period from 2017-2020. As a result, each 24 × 24 pixels S2 tile consists of 221 bands (17 *dates* × 13 *bands*).

The ALS data was collected in 2017 at the end of the dry season (May) by The National Centre for Airborne Laser Mapping (NCALM). NCALM also pre-processed the data, converting it from full-waveform to point cloud data and performing a point-wise classification into ground/non-ground classes. It must be noted that an additional ground classification have been performed by ZRC SAZU[2] to further increase its quality. Finally, among the most traditional ways of representing the DEMs derived from ALS for archaeological applications, we find the sky-view factor (SVF) [24], openness plots [6] and slope maps [16]. It must be noted that, these visual representations are used to highlight the natural and manmade features of the terrain. In this data-set, each 480 × 480 pixels ALS tiles comprises the sky-view factor, positive openness, and slope bands.

Resulting tiles have been labelled providing ground truth masks for three classes: buildings, platforms and aguadas. Buildings and platforms have been annotated through visualization for archaeological topography (VAT) [13] and a locally stretched elevation model, whereas aguadas have been annotated using the local dominance.

Overall, the data-set comprises 2,094 tiles, splited into training (1,765 tiles about an 82%) and testing (329 tiles about an 18%) sets. Since the testing set was used for leaderboard calculations, it has been disregarded for this study. Figure 4.2 presents a tile from the data-set.

### 4.2.2 Performance measures and evaluation

The performance has been quantitatively evaluated using the Intersection over Union (IoU) [17] metric. The IoU measures the overlap between the predicted segmentation and ground truth masks, divided by the area of union between both masks. Accordingly, the lower the IoU, the worse the prediction result. In this challenge, the overall average IoU score has been used to determine the winning solutions.

In this work, the IoU has been computed for both positive (building, platform or aguada) and negative (background) classes. Moreover, the mean of these IoU results has been also calculated. Apart from the aforementioned metrics, additional statistical measures such as the true positive rate (TPR), false positive rate (FPR), true negative rate (TNR), false negative rate (FNR) and positive predicted value (PPV) have been included. Finally, results corresponding to the 5 folds of each experiment have been averaged.

---

[2] https://www.zrc-sazu.si/

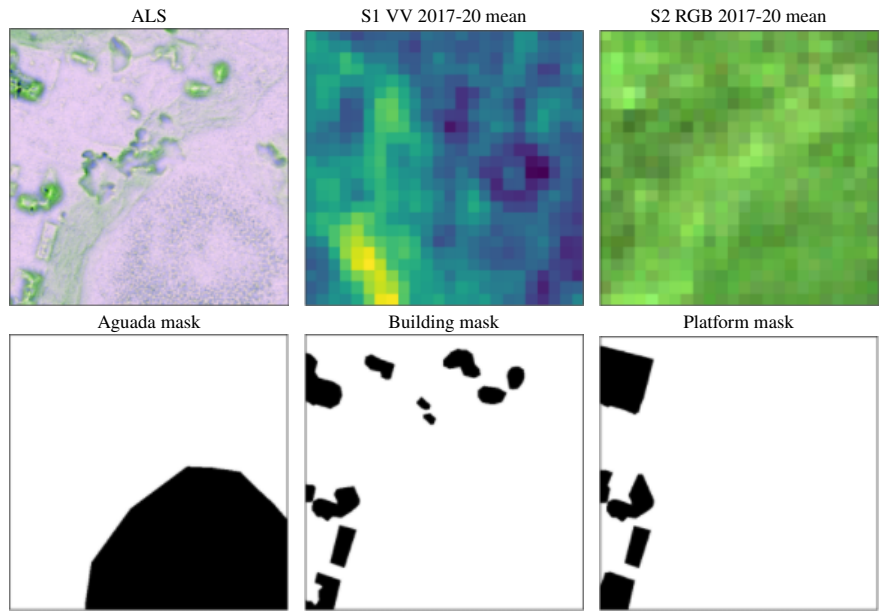

**Fig. 4.2** Close look at a tile corresponding to the challenge's training set.

## 4.3 Proposed solution

Despite the wealth of data provided in this challenge, we finally opted for only making use of the ALS data, on account of the limited resolution provided by S1 and S2 satellites and the results obtained in preliminary experiments.

The detection of buildings, platforms and aguadas in ALS data can be seen as a multi-class semantic segmentation problem. There is no prior certainty about how a multi-class semantic segmentation problem should be properly tackled. Usually, we just assume that a direct multi-class semantic segmentation approach would suffice. However, since this approach optimizes the overall performance, the accuracy over each class highly depends on their separability. In complex scenarios such as the location and identification of "lost" ancient Maya settlements hidden under the thick forest canopy, binary decomposition strategies [8] may come in handy. In this regard, our approach mainly consists on decomposing the multi-class semantic segmentation problem into multiple (a priori simpler) binary ones.

To decompose the multi-class problem, the One-vs-All (OVA) [1] binary decomposition strategy has been applied. Accordingly, the 3-classes semantic segmentation problem has been decomposed into three binary semantic segmentation ones (one for each class). For each binary semantic segmentation sub-problem, a different deep learning approach has been considered. However, all of them have some configurations in common.

A 5-fold set up has been considered to increase the generalization capabilities of the models. The way this split has been done varies depending on each class distribution. It should be noted that, as a result 5 different models have been be trained for each target class, which are used as an ensemble for predicting the test set. Moreover, the U-Net architecture [18] has been chosen as the main semantic segmentation architecture in all cases. However, the feature extractor (backbone) has been specialized depending on the scenario. It must be noted that all the backbones considered are Residual Networks [10]. Models have been trained for 100 epochs taking batches of 24 samples. The spatial dimensions of these samples (height and width) varies depending on the scenario. In order to take advantage of the strengths of a distribution-based loss and a region-based loss, a combination between the Cross-Entropy [15] and Log-Cosh Dice [11] losses has been considered. Moreover, in this work we have opted for the state-of-the-art One-Cycle-Policy [19] learning rate scheduler devoting a 10% of the cycle to the annealing of the learning rate. Furthermore, the AdamW [14] optimizer has been chosen, with a maximum learning rate of 1e-3 and weight decay of 1e-4. Additionally, dihedral [5] data augmentations (combinations of 90 degree rotations and flips) have been applied not only at training time but also when inferring. In the following, we describe the specific configurations for each class.

#### 4.3.0.1 Buildings.

The building class was the one with the greatest number of annotated pixels. In this case, the 5-fold split has been done stratifying depending on whether the building masks contain annotated pixels or not. Regarding the backbone, the ResNet-34 has been chosen, which has been pretrained on Imagenet [4]. In this particular scenario, samples of $256 \times 256$ have been considered. This samples have been randomly taken from the $480 \times 480$ ALS tiles.

#### 4.3.0.2 Platforms.

The platform class was the second one with the lowest number of annotated pixels. In this case, the 5-fold split has been carried out in the same way as in Buildings, depending on whether the platform masks contain annotated pixels or not. Regarding the backbone, the ResNet-101_32x4d has been chosen, which has been pretrained on Semi-Weakly Supervised ImageNet (SWSL) [23]. In this particular scenario, samples of $256 \times 256$ have been considered. This samples have been randomly taken from the $480 \times 480$ ALS tiles given a set of pre-computed pixel coordinates. That is, only pixel coordinates from which $256 \times 256$ samples can be taken having more than a 0.005 percent of the pixels annotated are considered.

#### 4.3.0.3 Aguadas.

The aguada class was by far the one with the lowest number of annotated pixels. In this case, the 5-fold split has been done stratifying depending on the number of annotated pixels the aguada masks have. That is, the percentage of annotated pixels have been computed and discretized into three categories using the following thresholds: 0, 0.05 and 0.15 (see Figure 4.3). Then, the folds are stratified using these categories. Regarding the backbone, the ResNet-34 has been chosen, which has been pretrained on Imagenet [4]. In this particular scenario, samples of $352 \times 352$ have been considered, since aguadas are bigger elements than buildings or platforms. This samples have been randomly taken from the $480 \times 480$ ALS tiles.

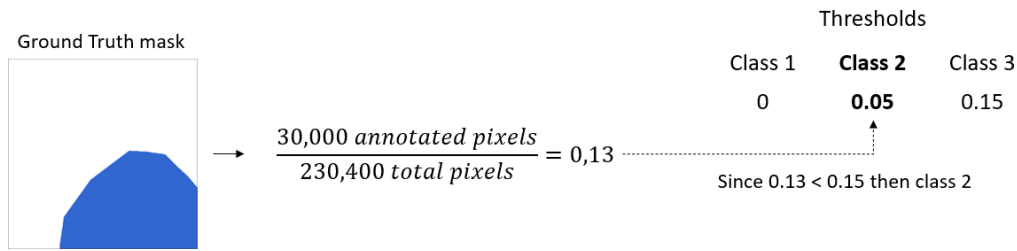

**Fig. 4.3** Visual example of the discretization process of the percentage of annotated pixels.

### 4.4 Results

In this section the results corresponding to the proposed solution described in Section 4.3 are discussed. Table 4.1 presents the results in terms of averaged IoU and several statistical measures. Additionally, the scores achieved in the final leader board (LB) in terms of IoU are also provided.

| Class | Set | IoU+ | IoU− | mIoU | TPR | FPR | TNR | FNR | PPV | LB | 1st |
|---|---|---|---|---|---|---|---|---|---|---|---|
| Building | Train | 0.7330 | 0.9948 | 0.8639 | 0.8454 | 0.0023 | 0.9976 | 0.1545 | 0.8639 | 0.7386 | 0.7530 |
|  | Valid | 0.7215 | 0.9946 | 0.8581 | 0.8269 | 0.0024 | 0.9976 | 0.1730 | 0.8560 |  |  |
| Platform | Train | 0.7367 | 0.9899 | 0.8633 | 0.8619 | 0.0053 | 0.9946 | 0.1380 | 0.8447 | 0.7082 | 0.7651 |
|  | Valid | 0.6350 | 0.9899 | 0.8125 | 0.8201 | 0.0061 | 0.9938 | 0.1798 | 0.7464 |  |  |
| Aguada | Train | 0.3815 | 0.9969 | 0.6892 | 0.7285 | 0.0017 | 0.9982 | 0.2714 | 0.6692 | 0.9863 | 0.9844 |
|  | Valid | 0.4034 | 0.9979 | 0.7006 | 0.6785 | 0.0010 | 0.9989 | 0.3214 | 0.6877 |  |  |
| Average | Train | 0.6171 | 0.9939 | 0.8055 | 0.8119 | 0.0031 | 0.9968 | 0.1880 | 0.7926 | 0.8110 | 0.7530 |
|  | Valid | 0.5866 | 0.9941 | 0.7904 | 0.7752 | 0.0032 | 0.9968 | 0.2247 | 0.7634 |  |  |

**Table 4.1** Averaged results across the 5 folds obtained for each target class.

Results highlight the remarkable capabilities of the models to detect buildings, platforms and aguadas (0.7386, 0.7082 and 0.9863 of IoU in the LB, respectively). Overall, these three models scored an mIoU across all classes of 0.8110, securing the 7th position in the "Discover the mysteries of the Maya" Challenge. However, these results are slightly far from the first position (0.7530, 0.7651 and 0.9844 of IoU in the LB for the building, platform and aguada classes, respectively).

### 4.5 Conclusions and future work

This work presents a deep learning approach to tackle the detection of ancient Maya architectures using ALS data. Taking into account the complexity of the multi-class semantic segmentation scenario, an OVA binary decomposition strategy has been used. This outlook allows to specify approaches for each target class. Finally, the results obtained show the outstanding capabilities of these models to detect buildings, platforms and aguadas. It must be noted that, this approach scored an overall IoU of 0.8110 in the "Discover the mysteries of the Maya" Challenge, which corresponds to the 7th place.

Nevertheless, results can be further improved making use of the multi-temporal high-resolution satellite imagery provided. Additionally, advanced pre-processing techniques may be applied to ALS data to better exploit this information. Finally, post-processing steps such as hole filling algorithms or Conditional Random Fields (CRFs) might be used to enhance the quality of the resulting mappings.

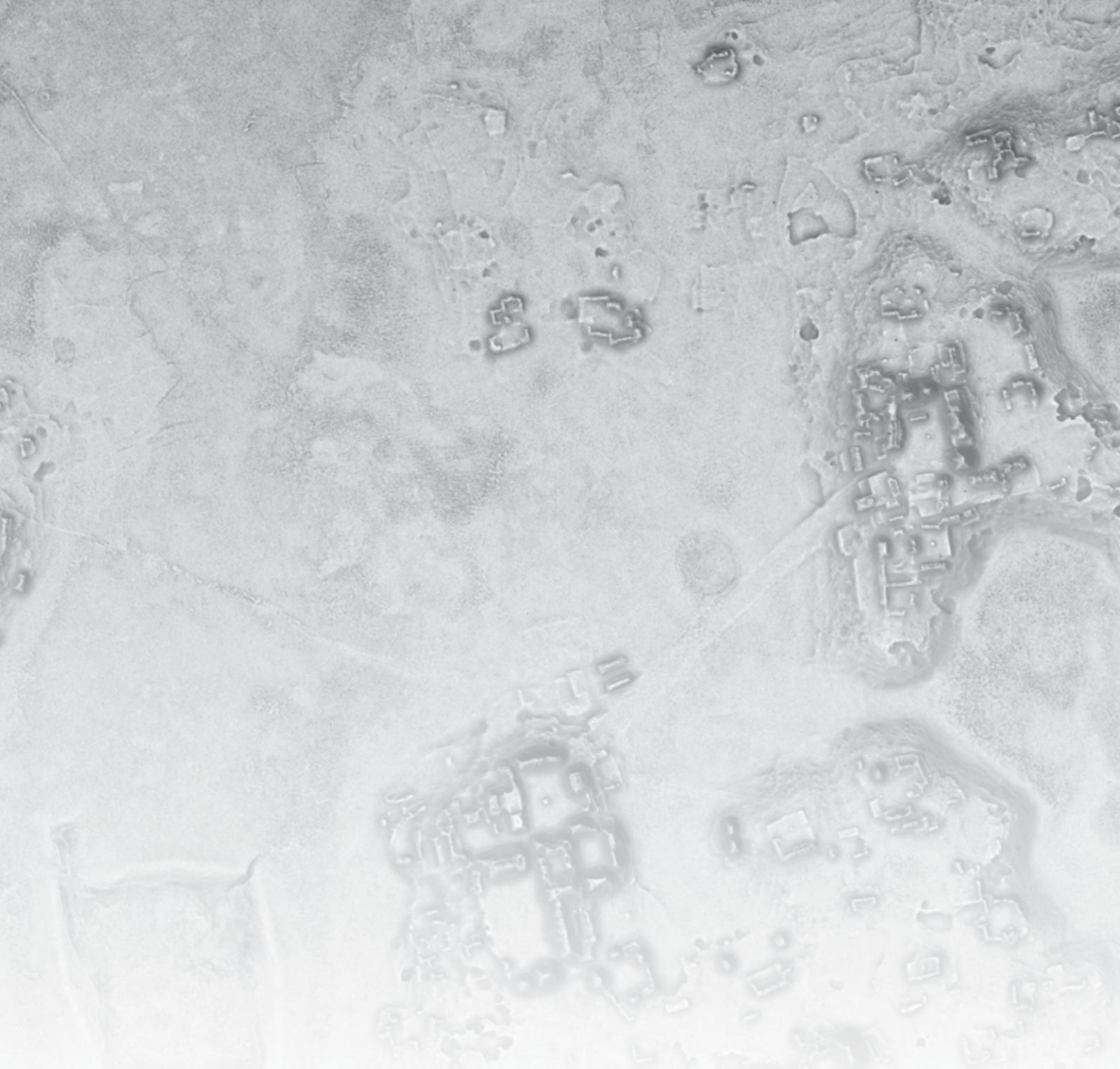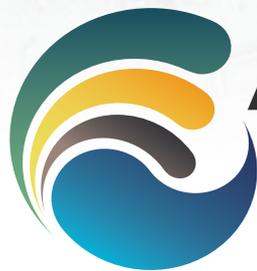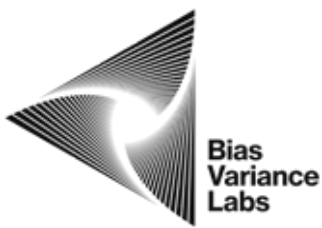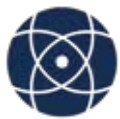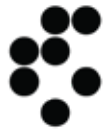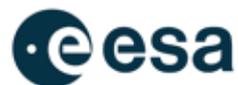